\documentclass[11pt, oneside]{article}
\usepackage[utf8]{inputenc}
\usepackage[margin=1in]{geometry}
\usepackage{float}
\usepackage{authblk}
\usepackage{amssymb, amsmath, amsthm}

\usepackage{mathtools}
\usepackage{algorithm}
\usepackage[noend]{algpseudocode}
\usepackage{todonotes}
\usepackage{graphicx}
\usepackage{hyperref}
\usepackage[backend=biber, style=numeric-comp, url=false,
            doi=false, isbn=false, eprint=false, natbib=true,
            sorting=none, uniquelist=false]{biblatex}
\AtEveryBibitem{\clearfield{month}}
\AtEveryBibitem{\clearfield{day}}
\AtEveryBibitem{\clearfield{version}}

\title{Mixtures of Gaussian process experts based on kernel stick-breaking processes}
\author{Yuji Saikai\thanks{ysaikai@unimelb.edu.au} }
\author{Khue-Dung Dang}
\affil{School of Mathematics and Statistics, the University of Melbourne}
\date{}
\begin{document}

\maketitle

\begin{abstract}
\noindent
Mixtures of Gaussian process experts is a class of models that can simultaneously address two of the key limitations inherent in standard Gaussian processes: scalability and predictive performance. In particular, models that use Dirichlet processes as gating functions permit straightforward interpretation and automatic selection of the number of experts in a mixture. While the existing models are intuitive and capable of capturing non-stationarity, multi-modality and heteroskedasticity, the simplicity of their gating functions may limit the predictive performance when applied to complex data-generating processes. Capitalising on the recent advancement in the dependent Dirichlet processes literature, we propose a new mixture model of Gaussian process experts based on kernel stick-breaking processes. Our model maintains the intuitive appeal yet improve the performance of the existing models. To make it practical, we design a sampler for posterior computation based on the slice sampling. The model behaviour and improved predictive performance are demonstrated in experiments using six datasets.

\vspace{10pt}
\end{abstract}



\vspace{10pt}
\section{Introduction}
The Gaussian process (GP) is a stochastic process and commonly used as a non-parametric Bayesian model that assumes a prior distribution over smooth functions. As a result, when predicting a response variable, GPs return not only mean prediction but also its variability \citep{rasmussen2006gaussian}. It has been used in a wide range of problems including spatial statistics \citep{cressie_statistics_1993}, emulation of computer experiments \citep{santner_design_2003}, and Bayesian optimisation \citep{shahriari_taking_2015}.

Standard GPs, which are based on stationary covariance functions and Gaussian homoskedastic noise, have two known limitations: scalability and predictive performance. First, the computational complexity of GPs is dominated by the computation of determinant and inverse of covariance matrices and scales as \(O(N^3)\) where \(N\) is the number of data points. As a result, standard GPs are often impractical for large \(N\). A variety of techniques have been developed to improve the scalability of standard GPs \citep{liu2020gaussian}. Second, by construction, standard GPs may struggle with non-stationary datasets \citep{nott_estimation_2002,schmidt_bayesian_2003,snoek_input_2014}. For example, in geo-statistical applications where the response variable is highly dependent on categorical features of locations, we expect sharp transitions in covariance between two locations characterised by two distinct categorical features \citep{kim_analyzing_2005}. Moreover, modelling heteroskedastic and/or multi-modal noise distributions requires extension of standard GPs \citep{wang2012gaussian}.

Mixtures of GP experts \citep{rasmussen_infinite_2001} are intuitive yet effective models that overcome both limitations of standard GPs. In general, the mixture of experts \citep{jacobs1991adaptive} is a class of models that probabilistically partition the input space and assign a separate predictive model to each of the local regions. Each predictive model responsible for an assigned local region is called an expert, and the assignment device is called a gating function. A mixture of GP experts is a model that uses a GP as an expert and has been used in a wide range of applications \citep{liu2020gaussian}.

Since each GP expert makes prediction based on only the data points contained in the assigned region, if expert assignment is deterministic, the computational complexity in prediction based on a subset of the data is clearly no greater and can be much lower than those based on all the data points. For example, if \(M\) experts are assigned to the equal number of data points (i.e., \(N/M\) data points per expert), the computational complexity is \(O(N^3/M^3)\). Moreover, even if expert assignment is stochastic and all experts are always assigned with positive probabilities, the overall complexity is still \(O(N^3/M^2)\).

Predictive performance can be significantly increased by mixtures of GP experts. First of all, as evident in basic Gaussian mixture models, multi-modal distributions naturally arise when multiple experts are mixed. In addition, non-trivial gating functions create different mixtures for different inputs, resulting noise distribution is automatically heteroskedastic. Finally, for the same reason, even if all the component GPs are based on stationary covariance functions, resulting covariance functions are non-stationary. While there exist a number of simple methods to effectively address the first two aspects, directly designing non-stationary covariance functions are known to be challenging \citep{fuglstad_does_2015}. The existing methods include input-space warping \citep{schmidt_bayesian_2003, anderes2008estimating, snoek_input_2014} and process convolution \citep{paciorek2006spatial}. Therefore, the third aspect is a significant advantage of mixtures of GP experts, which lead to non-stationary models without extra effort.

As is the case in finite mixture models, manually choosing a number of experts may be difficult if modellers possess little knowledge about the underlying processes they try to model. In this regard, as pointed out in \citep{rasmussen_infinite_2001}, use of the Dirichlet process (DP) is advantageous because it automatically chooses the number of experts through posterior sampling. Specifically, a gating function that automatically determines the number of experts can be constructed based on an input-dependent DP. \citet{rasmussen_infinite_2001} construct such a gating function by heuristically modifying the occupation number, which appears in the input-independent DP, into an input-dependent proportion of the total number of data points.

Although this is a simple and intuitive way to construct a dependent DP, such simplicity may also limit the capacity of resulting gating functions, which is crucial for predictive performance of mixture-of-experts models. Over the past two decades after the publication of \citep{rasmussen_infinite_2001}, the literature on dependent DPs and Bayesian nonparametrics in general have considerably grown and proposed many advanced methods \citep{quintana_dependent_2022}. Among them is kernel stick-breaking processes (KSBPs) \citep{dunson_kernel_2008}, which share with \citet{rasmussen_infinite_2001} an intuitive notion of similarity between two inputs captured by kernels. For the purpose of designing gating functions, thus, KSBPs allow us to maintain the intuitive appeal of the original gating function yet improve it to more capable ones for increased predictive performance.

In this paper, we construct a mixture of GP experts model in a principled way and improve the original model proposed by \citet{rasmussen_infinite_2001} in terms of predictive performance. Despite the improved performance, we keep the model intuitive and interpretable by providing straightforward interpretation of each parameter. The model is made practical by the accompanied sampler for posterior computation. In particular, we design a simpler algorithm for KSBP using the slice sampling against the retrospective MCMC adopted in the original work \citep{dunson_kernel_2008}.

First, we describe the model, a mixture of GP experts based on KSBP in Section \ref{sec:model}. Next, we explain in detail the posterior sampling in Section \ref{sec:sampler}. Then, in Section \ref{sec:experiments}, we conduct experiments both for illustration of the model behaviour (Part 1) and for performance comparison using five different datasets generated by five test functions found in the computer simulation literature, where GPs are commonly used as emulators (Part 2). Finally, in Section \ref{sec:discussion}, we discuss the model, possible extensions, and existing challenges to overcome.

\section{Model}\label{sec:model}
Suppose a regression problem with data \(\{(x_n,y_n)\}_{n=1}^N\) where \(x_n \in \mathbb{R}^D\) and \(y_n \in \mathbb{R}\). Then, the following is the model that consists of an infinite number of GP experts, each of which is responsible for a subset of data points probabilistically assigned by the gating function. By marginalising out the expert assignments \(\mathbf{s}\), the likelihood is found as follows:
\begin{align*}
p(\mathbf{y}|\mathbf{x},\theta)
  &= \sum_\mathbf{s} p(\mathbf{s}|\mathbf{x},\eta) \prod_i p(\mathbf{y}_i|\mathbf{s}_i,\mathbf{x}_i,\theta_i)\\
  &= \sum_\mathbf{s} \prod_n p(s_n|x_n,\eta) \prod_i p(\mathbf{y}_i|\mathbf{s}_i,\mathbf{x}_i,\theta_i),
\end{align*}
where \(\theta_i\) is a set of hyper-parameters that characterise the \(i\)th GP expert \((i\in\{1,2,\dots\})\), \(\mathbf{s} = \{s_n\}_n^N\) is the expert assignments \((s_n\in\{1,2,\dots\})\) for all \(n\), and \(p(s|x,\eta)\) is the gating function parameterised by \(\eta\). Under the KSBP gating function, for all \(n\in\{1,\dots,N\}\) and \(i\in\{1,2,\dots\}\), the input-dependent mixture weight is defined as follows:
\begin{align*}
w_{n,i} &= p(s_n=i|x_n,\eta)\\
        &= v_i \kappa(x_n,h_i)\left(1-\sum_{j=1}^{i-1}w_{n,j}\right)\\
        &= v_i \exp\left(-\left\|\frac{x_n-h_i}{r}\right\|^2\right) \left(1-\sum_{j=1}^{i-1}w_{n,j}\right),
\end{align*}
where \(v_i\) is the stick-breaking probability, \(h_i\) is the location of the \(i\)th expert, and \(\kappa(x,h)\) is the kernel based on scaled Euclidean distance between \(x\) and \(h\).

To complete the model, we specify the prior GPs using the constant mean function of \(0\) and the covariance function \(k_i\) such that
\begin{align*}
k_i(x_m,x_n) &= \sigma_i^2c_i(x_m,x_n) + \delta_{m,n}\tau_i^2\\
c_i(x_m,x_n) &= \exp\left(-\sum_{d=1}^D\left(\frac{x_{m,d}-x_{n,d}}{l_{i,d}}\right)^2\right),
\end{align*}
for all \(i\in\{1,2,\dots\}\). In this model, we use one of the most basic covariance functions---the squared exponential covariance function with length-scale parameter \(l_{i,d}\) for each dimension \(d\in\{1,\dots,D\}\) and independent additive noise, which is implied by the Kronecker delta \(\delta_{m,n}\) in front of \(\tau_i^2\).

\begin{table}[H]
\centering
\caption{Key notations.}
\label{table:notations}
\vspace{.3cm}
\begin{tabular}{ l|l }
\(s_n\) & Expert assignment for the \(n\)th data point\\\hline
\(k_i(x_m,x_n)\) & \(i\)th covariance function between \(x_m\) and \(x_n\)\\\hline
\(c_i(x_m,x_n)\) & \(i\)th correlation function between \(x_m\) and \(x_n\)\\\hline
\(\theta_i\) & \((\sigma_i^2,l_i,\tau_i^2)\), hyper-parameters for the \(i\)th GP\\\hline
\(\sigma_i^2\) & \(i\)th output-scale\\\hline
\(l_i\) & \(i\)th length-scale\\\hline
\(\tau_i^2\) & \(i\)th noise variance\\\hline
\(v_i\) & \(i\)th stick-breaking probability\\\hline
\(h_i\) & Location of the \(i\)th expert in the input space\\\hline
\(\kappa(x,h)\) & Kernel returning a scaled distance between \(x\) and \(h\)\\\hline
\(r\) & Kernel width for scaling
\end{tabular}
\end{table}

We take a fully Bayesian approach to model learning and assume the following classes of priors. For \(\sigma_i^2\), \(l_i\), \(\tau_i^2\) and \(r\), we use gamma distributions. For \(h_i\), we use the uniform distribution over the \(D\)-dimensional input space, as it reflects the lack of prior knowledge about expert's locations. By construction, we have \(v_i\sim\text{beta}(\alpha, \beta)\) where \(\alpha=1\) corresponds to standard DPs. We use geometric distributions for \(\alpha\) and \(\beta\). The reason for the geometric distribution is twofold. First, the prior for \(v_i\) is updated based on auxiliary Bernoulli trials. Therefore, the use of discrete distribution over positive integers is natural in specifying prior ``success'' and ``failure'' counts. Second, the functional form of geometric distribution simplifies the posterior mass functions and helps us design samplers based on rejection sampling. Both points are expanded on in the next section.

\section{Posterior sampling}\label{sec:sampler}
While use of KSBPs is a promising approach to designing gating functions, there is one aspect in posterior sampling that can be simplified without losing its advantage. Recall that we need to truncate the infinite sequence of stick-breaking process to operationalise it into finite computation. To this end, \citet{dunson_kernel_2008} use the retrospective MCMC \citep{papaspiliopoulos_retrospective_2008}, a complex algorithm that involves Metropolis-Hastings steps. For our purpose, such complex machinery is unnecessary, and random truncation can be achieved by the slice sampling \citep{walker_sampling_2007}, which makes posterior sampling simpler and computationally lighter \citep{kalli_slice_2011}. In what follows, thus, we take this approach in designing the sampler.

Besides the model parameters, the following auxiliary variables and parameters are also sampled at each round of the posterior sampling:

\begin{table}[H]
\centering
\begin{tabular}{ l|l }
\(u_n\) & Slicing variable for the \(n\)th data point \((u_n\in[0,1])\)\\\hline
\(A_{n,i}\) & Bernoulli indicator such that \(p(A_{n,i}=1) = v_i\)\\\hline
\(B_{n,i}\) & Bernoulli indicator such that \(p(B_{n,i}=1) = \kappa(x_n,h_i)\)
\end{tabular}
\end{table}

\noindent The slicing variables \(\{u_n\}\) is a set of random truncation points and control the sufficient numbers of candidate experts, which play a role in sampling \(v\), \(h\) and \(\mathbf{s}\). To interpret \(A_{n,i}\) and \(B_{n,i}\), note that the KSBP gating function uses two Bernoulli trials when probabilistically assigning the \(n\)th data point to the \(i\)th expert: one based on \(v_i\) and the other based on \(\kappa(x_n,h_i)\), which depends on \(h_i\) and \(r\). Thus, \(\{A_{n,i}\}_n^N\) provides the `success' counts that define the likelihood for \(v_i\). Similarly, \(\{B_{n,i}\}_n^N\) provides the `success' counts that define the likelihood for \(h_i\), and \(\{B_{n,i}\}_n^N\) for all \(i\) together provides the `success' counts that define the likelihood for \(r\).

We design a within-Gibbs sampler to learn the model described in Section \ref{sec:model}. In each iteration of Gibbs sampling, blocks of parameters are sequentially updated by sampling from their conditional posteriors through direct sampling due to the conjugacy, rejection sampling \citep{gilks1992adaptive}, or Hamiltonian Monte Carlo (HMC) \citep{neal2011mcmc}. The specifications of HMC are provided in the appendix. Algorithm \ref{alg:sampler} outlines the sampling steps, where \(M\) is the total number MCMC iterations and \(weight\) is the gating function to compute \(p(s_n=i|x_n,\eta)\) for all \(n\) and \(i\).

{
\centering
\begin{minipage}{.53\linewidth}
\begin{algorithm}[H]
\caption{Within-Gibbs sampler for GPKSBP}\label{alg:sampler}
\begin{algorithmic}[1]
  \State \textbf{require} \(M, weight\)
  \State Initialise \(r,v,h,\alpha,\beta,\mathbf{s},\theta\)
  \For {\(m\in\{1,2,\dots,M\}\)}
    \State Sample \(r\) via HMC
    \State \(u_n \gets 0\) for all \(n\)
    \State \(j \gets 0\)
    \While {\(u_n \leq 1 - \sum_{j} w_{n,j}\) for some \(n\)}
      \State \(j \gets j+1\)
      \State Sample \(A_{n,j}\) and \(B_{n,j}\) for all \(n\)
      \State Sample \(v_{j}\)
      \State Sample \(h_{j}\) via HMC
      \State \(w_{n,j} \gets weight(x_n,j)\) for all \(n\)
      \State Sample \(u_n\) for \(n\) such that \(s_n=j\)
    \EndWhile
    \State \(i^* \gets j\)
    \State Sample \(\alpha\) and \(\beta\) via rejection sampling
    \State Sample \(s_n\) for all \(n\)
    \State Sample \(\theta_i\) for \(i\in\{1,2,\dots,i^*\}\) via HMC
  \EndFor
\end{algorithmic}
\end{algorithm}
\end{minipage}
\par
\vspace{1em}
}

\noindent Section 3.1 to 3.5 explain in detail each of the steps in one MCMC iteration. In particular, Line 5 to 14 correspond to Section \ref{sec:vhu} for sampling \(v\), \(h\) and \(u\).

\subsection{\(r\)}
The conditional posterior of \(r\) is derived through the likelihood of \(\{\{B_{n,i}\}_{s_n\geq i}\}_{i=1}^{i^*}\), which is parameterised by \(\{\{\kappa_{n,i}\}_{s_n\geq i}\}_{i=1}^{i^*}\), which is in turn a function of \(r\). Thus,
\begin{align*}
p(r | \{\{B_{n,i}\}_{s_n\geq i}\}_{i=1}^{i^*})
&\propto p(r, \{\{B_{n,i}\}_{s_n\geq i}\}_{i=1}^{i^*})\\
&= p(r) \prod_{i=1}^{i^*} p(\{B_{n,i}\}_{s_n\geq i} | r)\\
&= p(r) \prod_{i=1}^{i^*} \prod_{s_n\geq i} \kappa_{n,i}^{B_{n,i}}(1-\kappa_{n,i})^{1-B_{n,i}}.
\end{align*}
where each \(\kappa_{n,i}\) is a function of \(r\), and \(i^*\) is the sufficient number of candidate experts. \(i^*\) and the justification of
\[
p(\{B_{n,i}\}_{s_n\geq i} | r) = \prod_{s_n\geq i} \kappa_{n,i}^{B_{n,i}}(1-\kappa_{n,i})^{1-B_{n,i}}
\]
are explained in the next subsection. To sample from this density, we use HMC.

\subsection{\(v\), \(h\) and \(u\)}\label{sec:vhu}
Following \citet{dunson_kernel_2008}, we alternately sample \((v_i, h_i)\) and auxiliary \((A_{n,i}, B_{n,i})\) for each \(i\) starting with \(i=1\) until the stopping criterion is met. The stopping criterion is derived from the one used for input-independent DPs \citep{walker_sampling_2007}. In the slice sampling for input-independent DPs, it is sufficient to generate \(i^*\) number of weights where \(i^*\) is the smallest positive integer that satisfies
\[
\min\{u_n\} > 1 - \sum_{i=1}^{i^*} w_{i}.
\]
The idea behind is the following. The right-hand side is the remaining weight, and \(w_i\) for all \(i>i^*\) cannot exceed it. To ensure \(u_n > 1-\sum w_i\) for all \(n\), it uses a conservative condition, \(\min\{u_n\} > 1-\sum w_i\).  In our model, however, weights \(\{w_{n,i}\}\) are input-dependent, so a modified sufficient condition is
\[
u_n > 1 - \sum_{i=1}^{i^*} w_{n,i}
\]
for all \(n\). In other words, we stop sampling \((v_i, h_i)\) after \(i=i^*\) for the first time.

For each \(i\), we first sample \(A_{n,i}\) and \(B_{n,i}\) for all \(n\) given \(v_i\) and \(\kappa_{n,i}\). Since \(A_{n,i}\) and \(B_{n,i}\) are Bernoulli random variables with success probability \(v_i\) and \(\kappa_{n,i}\) respectively, given \(s_n\), we have the following distribution of \((A_{n,i}, B_{n,i})\). For \(n\) such that \(s_n=i\),
\[
p(A_{n,i}, B_{n,i}) =
  \begin{cases}
    1 & \text{for } (1,1)\\
    0 & \text{otherwise}
  \end{cases}.
\]
For \(n\) such that \(s_n>i\),
\[
p(A_{n,i}, B_{n,i}) \propto
  \begin{cases}
    v_i(1-\kappa_{n,i}) & \text{for } (1,0)\\
    (1-v_i)\kappa_{n,i} & \text{for } (0,1)\\
    (1-v_i)(1-\kappa_{n,i}) & \text{for } (0,0)\\
    0 & \text{for } (1,1)
  \end{cases}.
\]
For the reason explained below, we need not sample for \(n\) such that \(s_n<i\).

For \(v_i\), due to the conjugacy, given \(\{A_{n,i}\}_{n=1}^N\), the posterior of \(v_i\) is also the beta distribution:
\[
v_i \sim \text{beta}\Bigg(\alpha+\sum_{n: s_n\geq i}A_{n,i},\; \beta+\sum_{n: s_n\geq i}(1-A_{n,i})\Bigg).
\]
Note that to update \(v_i\), the relevant likelihood consists of only \(A_{n,i}\) such that \(s_n\geq i\) because, first, the assignments to the \(j\)th experts where \(j<i\) are not influenced by the \(i\)th stick-breaking trials and, second, the \(i\)th stick-breaking trials precede the assignments to the \(j\)th experts where \(j>i\).

For \(h_i\), given \(\{B_{n,i}\}_{s_n\geq i}\), the posterior density is derived as follows.
\begin{align*} 
p(h_i | \{B_{n,i}\}_{s_n\geq i})
&\propto p(\{B_{n,i}\}_{s_n\geq i}, \{\kappa_{n,i}\}_{s_n\geq i}, h_i)\\
&= p(\{B_{n,i}\}_{s_n\geq i} | \{\kappa_{n,i}\}_{s_n\geq i}) \cdot p(\{\kappa_{n,i}\}_{s_n\geq i} | h_i) \cdot p(h_i)\\
&= \prod_{s_n\geq i} \kappa_{n,i}^{B_{n,i}}(1-\kappa_{n,i})^{1-B_{n,i}}\cdot 1 \cdot 1\\
&= \prod_{s_n\geq i} \kappa_{n,i}^{B_{n,i}}(1-\kappa_{n,i})^{1-B_{n,i}},
\end{align*}
where we use the fact that \(\kappa_{n,i}\) is the deterministic function of \(h_i\), and the prior of \(h_i\) is uniform. To sample from this density, we use HMC. The gradient of the log-posterior density is the following.
\begin{align*}
\nabla \log p(h_i | \{B_{n,i}\}_{s_n\geq i})
&= \sum_{s_n\geq i} B_{n,i}\nabla \log\kappa_{n,i} + (1-B_{n,i})\nabla \log(1-\kappa_{n,i})\\
&= \sum_{s_n\geq i} B_{n,i}\frac{\nabla\kappa_{n,i}}{\kappa_{n,i}} - (1-B_{n,i})\frac{\nabla\kappa_{n,i}}{1-\kappa_{n,i}},
\end{align*}
where, for \(d\in\{1,\dots,D\}\),
\[
\frac{\partial\kappa_{n,i}}{\partial h_{i,d}} = \kappa_{n,i}\frac{2(x_{i,d}-h_{i,d})}{r^2}.
\]

Finally, given \(w_{n,s_n}\), the conditional posterior for \(u\) is simple:
\[
u_n \sim \text{Unif}(0, w_{n,s_n}).
\]
However, as pointed out by \citet{kalli_slice_2011}, the distribution depends on \(w_{n,s_n}\), which depends on \(\{v_i\}\) and \(\{h_i\}\), whose sampling is terminated based on \(i^*\), which circularly depends on \(u_n\) as described above. In other words, we cannot sample \(u_n\) given \(w_{n,s_n}\) for all \(n\) at once; rather, we must interweave the sampling of \(u_n\) with the sampling of \(\{v_i\}\) and \(\{h_i\}\) and proceed in tandem. Consequently, we sample \(u_n\) for \(n\) such that \(s_n=i\) after sampling \(v_i\) and \(h_i\) and before sampling \(v_{i+1}\) and \(h_{i+1}\).

\subsection{\(\alpha\) and \(\beta\)}
The conditional mass function for positive integers \(\alpha\) and \(\beta\) is
\begin{align*}
  p(\alpha,\beta | \{v_i\})
&\propto p(\alpha,\beta) p(\{v_i\}|\alpha,\beta)\\
&= p(\alpha,\beta) \prod_{i=1}^{i^*} \frac{\Gamma(\alpha+\beta)}{\Gamma(\alpha)\Gamma(\beta)}v_i^{\alpha-1}(1-v_i)^{\beta-1}\\
&= p(\alpha,\beta) \prod_{i=1}^{i^*} \frac{(\alpha+\beta-1)!}{(\alpha-1)!(\beta-1)!} v_i^{\alpha-1}(1-v_i)^{\beta-1},
\end{align*}
where \(\Gamma\) is the gamma function, and we used the fact \(\Gamma(n) = (n-1)!\) for positive integer \(n\). The independence between the prior geometric distributions implies
\[
p(\alpha,\beta) = p(\alpha)p(\beta) = (1-p_\alpha)^{\alpha-1}p_\alpha (1-p_\beta)^{\beta-1}p_\beta,
\]
where \(p_\alpha\) and \(p_\beta\) are the corresponding ``success'' parameters. Thus, the posterior is
\begin{align*}
  p(\alpha,\beta | \{v_i\})
&\propto (1-p_\alpha)^{\alpha-1} (1-p_\beta)^{\beta-1} \prod_{i=1}^{i^*} \frac{(\alpha+\beta-1)!}{(\alpha-1)!(\beta-1)!} v_i^{\alpha-1}(1-v_i)^{\beta-1}\\
&= \left(\frac{(\alpha+\beta-1)!}{(\alpha-1)!(\beta-1)!}\right)^{i^*} \left((1-p_\alpha)\prod v_i\right)^{\alpha-1} \left((1-p_\beta)\prod (1-v_i)\right)^{\beta-1}.
\end{align*}

For alternate sampling, the following are the conditional mass functions for \(\alpha\) and \(\beta\):
\begin{align*}
  p(\alpha|\beta,\{v_i\})
&\propto \bar{p}(\alpha|\beta,\{v_i\})  = \left(\frac{(\alpha+\beta-1)!}{(\alpha-1)!}\right)^{i^*} \left((1-p_\alpha)\prod v_i\right)^{\alpha-1}\\
  p(\beta|\alpha,\{v_i\})
&\propto \bar{p}(\beta|\alpha,\{v_i\}) = \left(\frac{(\alpha+\beta-1)!}{(\beta-1)!}\right)^{i^*} \left((1-p_\beta)\prod (1-v_i)\right)^{\beta-1}.
\end{align*}
To see how \(p(\alpha|\beta,\{v_i\})\) evolves as \(\alpha\) increases, we examine the ratio 
\[\frac{p(\alpha+1|\beta,\{v_i\})}{p(\alpha|\beta,\{v_i\})}.\]
Note that the ratio of the first factor is
\[
\left(\frac{(\alpha+\beta)!}{(\alpha)!}\right)^{i^*} \Big/ \left(\frac{(\alpha+\beta-1)!}{(\alpha-1)!}\right)^{i^*} = \left(\frac{\alpha+\beta}{\alpha}\right)^{i^*}\]
which is \((1+\beta)^{i^*}\) at \(\alpha=1\) and monotonically decreases towards 1. Since the other ratio \((1-p_\alpha)\prod v_i\) is constant and less than 1, we may conclude that the posterior is single-peaked. Specifically, the peak is at \(\alpha^*\) such that
\begin{gather*}
   \left(\frac{\alpha^*+\beta}{\alpha^*}\right)^{i^*}(1-p_\alpha)\prod v_i < 1\
\end{gather*}
is satisfied for the first time.

With this fact, we can construct an envelope for rejection sampling to sample from \(p(\alpha|\beta,\{v_i\}) \). Specifically, we may use one that goes flat up to \(\alpha^*\) and then slopes down at the constant decay rate, which is slower than the posterior's decay rate as indicated by decreasing \(\left(\frac{\alpha+\beta}{\alpha}\right)^{i^*}\). Formally, we may use the uniform distribution over \(\alpha\in\{1,\dots,\alpha^*\}\) and the geometric distribution truncated at \(\alpha^*+1\) for \(\alpha>\alpha^*\) with ``failure'' probability equal to the above threshold value, which makes a thicker tail than the posterior.

Let \(\phi_\alpha\) denote the failure probability:
\[
\phi_\alpha = \left(\frac{\alpha^*+\beta}{\alpha^*}\right)^{i^*}(1-p_\alpha)\prod v_i.
\]
Then, the proposal mass function \(q(\alpha)\) is indirectly specified through the envelope \(c\cdot q(\alpha)\) for \(\bar{p}(\alpha|\beta,\{v_i\})\):
\begin{align*}
c\cdot q(\alpha)
&=
\begin{cases}
  \bar{p}(\alpha^* |\beta,\{v_i\}) & \text{for } \alpha \in \{1,\dots, \alpha^*\}\\
  \bar{p}(\alpha^* |\beta,\{v_i\}) \left(\phi_\alpha\right)^{\alpha-\alpha^*} & \text{for } \alpha \in \{\alpha^*+1,\alpha^*+2,\dots\}
\end{cases}
\end{align*}
where the normalisation constant \(c\) is
\[
c = \sum_{\alpha=1}^\infty c\cdot q(\alpha) = \bar{p}(\alpha^* |\beta,\{v_i\}) \left(\alpha^*+\frac{\phi_\alpha}{1-\phi_\alpha}\right).
\]
We may sample \(\hat{\alpha}\) from \(q\) as follows. With probability
\[
\sum_{\alpha=1}^{\alpha^*}q(\alpha) = \sum_{\alpha=1}^{\alpha^*}\left(\alpha^*+\frac{\phi_\alpha}{1-\phi_\alpha}\right)^{-1} = \alpha^*\left(\alpha^*+\frac{\phi_\alpha}{1-\phi_\alpha}\right)^{-1},
\]
uniformly sample \(\hat{\alpha}\) from \(\{1,\dots, \alpha^*\}\). With probability
\[
1 - \sum_{\alpha=1}^{\alpha^*}q(\alpha),
\]
sample \(\hat{\alpha}\) from \(\{\alpha^*+1,\alpha^*+2,\dots\}\) using the conditional mass function
\begin{align*}
q(\alpha|\alpha>\alpha^*)
&= \frac{q(\alpha)}{1 - \sum_{\alpha=1}^{\alpha^*}q(\alpha)}\\
&= \frac{\left(\alpha^*+\frac{\phi_\alpha}{1-\phi_\alpha}\right)^{-1}\left(\phi_\alpha\right)^{\alpha-\alpha^*}}{1 - \alpha^*\left(\alpha^*+\frac{\phi_\alpha}{1-\phi_\alpha}\right)^{-1}}\\
&= \left(\phi_\alpha\right)^{\alpha-\alpha^*-1}(1-\phi_\alpha),
\end{align*}
which implies \(\alpha-\alpha^*|\alpha>\alpha^* \sim \text{geometric}(1-\phi_\alpha)\). Finally, given the proposal \(\hat{\alpha}\) drawn from \(q\), draw \(u \sim \text{Unif}(0, c\cdot q(\hat{\alpha}))\) and, if
\[
u \leq \bar{p}(\hat{\alpha }|\beta,\{v_i\}) = \left(\frac{(\hat{\alpha}+\beta-1)!}{(\hat{\alpha}-1)!}\right)^{i^*} \left((1-p_\alpha)\prod v_i\right)^{\hat{\alpha}-1},
\]
then accept \(\hat{\alpha}\) as a sample of \(\alpha\); otherwise, repeat the process.

By symmetry, we also have \(\beta^*\) such that \(\phi_\beta < 1\) where
\[
\phi_\beta = \left(\frac{\beta^*+\alpha}{\beta^*}\right)^{i^*}(1-p_\beta)\prod (1-v_i).
\]
So, in a similar way, we can sample \(\beta\) using rejection sampling.

\subsection{\(\mathbf{s}\)}
As a consequence of the slice sampling \citep{walker_sampling_2007}, given \(u_n\), the posterior mass function of \(s_n\) is
\[
p(s_n | \mathbf{y}_{s_n}^{(-n)},\theta_{s_n}) \propto p(y_n | \mathbf{y}_{s_n}^{(-n)},\theta_{s_n}),
\]
for \(s_n \in \{i : u_n < w_{n,i}\}\), and 0 otherwise. Note that \(p(y_n | \mathbf{y}_{s_n}^{(-n)},\theta_{s_n})\) is the likelihood specified by the posterior Gaussian process conditional on the other data points assigned to the \({s_n}\)th expert. In case there is no other data point assigned to the expert, the relevant likelihood is specified by the prior Gaussian process---a univariate Gaussian distribution of mean 0 and variance \(\sigma_{s_n}^2 + \tau_{s_n}^2\) \citep{rasmussen_infinite_2001}.

\subsection{\(\theta\)}
Recall that \(\theta_i = (\sigma_i^2, l_i, \tau_i^2)\) is a set of hyper-parameters that specify the \(i\)th GP expert. Given the expert assignments \(\mathbf{s}\) and the responses \(\mathbf{y}\), for all \(i\in\{1,2,\dots\}\), the posterior is,
\[
p(\theta_i | \mathbf{y}_i) \propto p(\theta_i) p(\mathbf{y}_i |\theta_i)
\]
where \(p(\mathbf{y}_i |\theta_i)\) is the likelihood specified by the \(i\)th prior Gaussian process. We use HMC to sample from the posterior.

First, let \(z\) denote the momentum. Then, we use the standard normal distribution:
\[
z \sim \mathcal{N}(0, I).
\]
To compute the partial derivatives of the log likelihood needed for the gradient of potential \(U(\theta_i)\), we use Eq. (5.9) in \citet{rasmussen2006gaussian}[p.114]. For example,
\[
\frac{\partial}{\partial\sigma_i^2} \log p(\mathbf{y}_i |\theta_i) = \frac{1}{2} \text{tr}\left((K_i^{-1}\mathbf{y}_i\mathbf{y}_i^TK_i^{-1} - K_i^{-1})\frac{\partial K_i}{\partial\sigma_i^2}\right).
\]
For the covariance matrix \(K_i\) formed by the squared exponential \(k_i\) defined above,
\begin{align*}
\frac{\partial K_i}{\partial \sigma_i^2} &= C_i\\
\frac{\partial K_i}{\partial l_{i,d}} &= \sigma_i^2\frac{\partial C_i}{\partial l_{i,d}} = \frac{\sigma_i^2C_i\odot\Delta_{i,d}}{l_{i,d}^3}\\
\frac{\partial K_i}{\partial \tau_i^2} &= I.
\end{align*}
\(C_i\odot\Delta_{i,d}\) is the element-wise product, \(C_i\) is the correlation matrix, \(l_{i,d}\) is the \(d\)th dimension of \(l_i\), and \(\Delta_{i,d}\) is the matrix of squared differences in the \(d\)th dimension of \(x\) (and \(l\)):
\[
\Delta_{i,d} =
\begin{bmatrix}
(x_{1,d} - x_{1,d})^2 & \dots & (x_{1,d} - x_{N_i,d})^2\\
\vdots & \ddots & \vdots\\
(x_{N_i,d} - x_{1,d})^2 & \dots & (x_{N_i,d} - x_{N_i,d})^2\\
\end{bmatrix}
\]
where \(N_i\) is the number of data points associated with the \(i\)th expert. For gamma priors, the partial derivative of the log prior takes a simple form. For example, for \(\sigma_i^2 \sim \text{gamma}(p_1,p_2)\),
\[
\frac{\partial \log p(\sigma_i^2)}{\partial \sigma_i^2} \propto \frac{\partial}{\partial \sigma_i^2}\left[ (p_1-1)\log \sigma_i^2 - \frac{\sigma_i^2}{p_2} \right] = \frac{(p_1-1)}{\sigma_i^2} - \frac{1}{p_2}.
\]
Now the gradient of the potential is
\[
\nabla U(\theta_i) = -\nabla \log p(\theta_i) - \nabla \log p(\mathbf{y}_i|\theta_i).
\]

\section{Experiments}\label{sec:experiments}
The section consists of two parts: demonstration of GPKSBP (our model) using an illustrative dataset (Part 1) and performance comparison between GPKSBP and RG (the baseline model from \citet{rasmussen_infinite_2001}) using five different datasets (Part 2). For reproducibility, the random seeds used for replication are explicitly mentioned, and the code used to implement the sampler and run the experiments is made available on the \href{https://github.com/ysaikai/GPKSBP}{GitHub repository}.

All the datasets were generated by test functions found in the computer simulation literature, where GPs are commonly used as emulators. While we could have used real-world datasets such as geo-statistical ones, for which GPs are also commonly used, data-generating processes captured by those datasets turned out not complex enough in terms of dimensionality and non-stationarity to contrast GPKSBP with RG. For example, despite the non-stationarity strongly expected in underlying spatial processes, use of non-stationary GPs is unnecessary in many cases \citep{fuglstad_does_2015}. The use of test functions allows us to construct an illustrative dataset in Part 1 and examine predictive performance of both models in capturing sufficiently complex processes in Part 2. For meaningful comparison, the priors and the posterior sampling procedure were kept almost identical between GPKSBP and RG.

In each experiment, the raw data was pre-processed so that each dimension of the input \(X\) was normalised to $[0,1]$, and the response \(Y\) was standardised to mean of 0 and standard deviation of 1. In posterior sampling, the number of MCMC samples was 20,000, and the first 10,000 samples were discarded as burn-in samples.

We used fairly generic specifications of the priors to avoid undue tuning of GPKSBP to the selected datasets. The priors were specified as follows:

\begin{table}[H]
\centering
\caption{Prior specifications for GPKSBP}
\label{table:GPKSBPpriors}
\vspace{.3cm}
\begin{tabular}{ l|l }
\(\sigma_i^2\) & \(\text{gamma}(2, 2)\)\\\hline
\(l_{i,d}\) & \(\text{gamma}(2, 0.5)\) for all \(d\in\{1,\dots,D\}\)\\\hline
\(\tau_i^2\) & \(\text{gamma}(2, 0.5)\)\\\hline
\(r\) & \(\text{gamma}(2, 0.5)\)\\\hline
\(h_i\) & uniform over \([0,1]^D\)\\\hline
\(v_i\) & \(\text{beta}(\alpha, \beta)\)\\\hline
\(\alpha\) & \(\text{geometric}(0.5)\)\\\hline
\(\beta\) & \(\text{geometric}(0.5)\)
\end{tabular}
\end{table}

\noindent Note that \(h_i\), \(v_i\), and \(\alpha\) are unique to GPKSBP. \(\text{gamma}(p_1,p_2)\) denotes the gamma distribution specified by shape parameter \(p_1\) and scale parameter \(p_2\). The same class of covariance functions was used for RG. The priors were specified as follows:

\begin{table}[H]
\centering
\caption{Prior specifications for RG}
\label{table:RGpriors}
\vspace{.3cm}
\begin{tabular}{ l|l }
\(\sigma_i^2\) & \(\text{gamma}(2, 2)\)\\\hline
\(l_{i,d}\) & \(\text{gamma}(2, 0.5)\) for all \(d\in\{1,\dots,D\}\)\\\hline
\(\tau_i^2\) & \(\text{gamma}(2, 0.5)\)\\\hline
\(r\) & \(\text{gamma}(2, 0.5)\)\\\hline
\(\beta\) & \(\text{gamma}(2,1)\)
\end{tabular}
\end{table}

\noindent The only difference was the one for \(\beta\sim \text{gamma}(2,1)\), instead of \(\beta\sim\text{geometric}(0.5)\). \citet{rasmussen_infinite_2001} do not provide the prior specification, and we could use \(\beta\sim\text{geometric}(0.5)\) for RG as well. However, \(\beta\sim \text{gamma}(2,1)\) was chosen because of the conjugacy created by use of an auxiliary variable \citep{escobar1995bayesian}; i.e., \(\beta\) remains gamma-distributed and therefore simple to sample in the posterior sampling. Note that the mean \((=2)\) and the variance \((=2)\) were matched between two priors of \(\beta\).

We supplemented a few details of RG sampler omitted in the original paper. For example, \citet{rasmussen_infinite_2001} use ``Neal's 8'' algorithm \citep{neal_markov_2000} to sample \(\mathbf{s}\) but do not provide information about the number auxiliary parameters. We used one auxiliary parameter. Another example is sampling \(r\), for which \citet{rasmussen_infinite_2001} use the Metropolis method with a Gaussian proposal to sample from the pseudo-posterior. Since the variance used for the Gaussian proposal is unknown, we used HMC instead to sample from the pseudo-posterior, which is documented in the appendix. Note that we also used HMC to sample \(r\) in GPKSBP.

\subsection{Part 1. Model demonstration}
\begin{figure}[H]
    \centering
    \includegraphics[width=.6\linewidth]{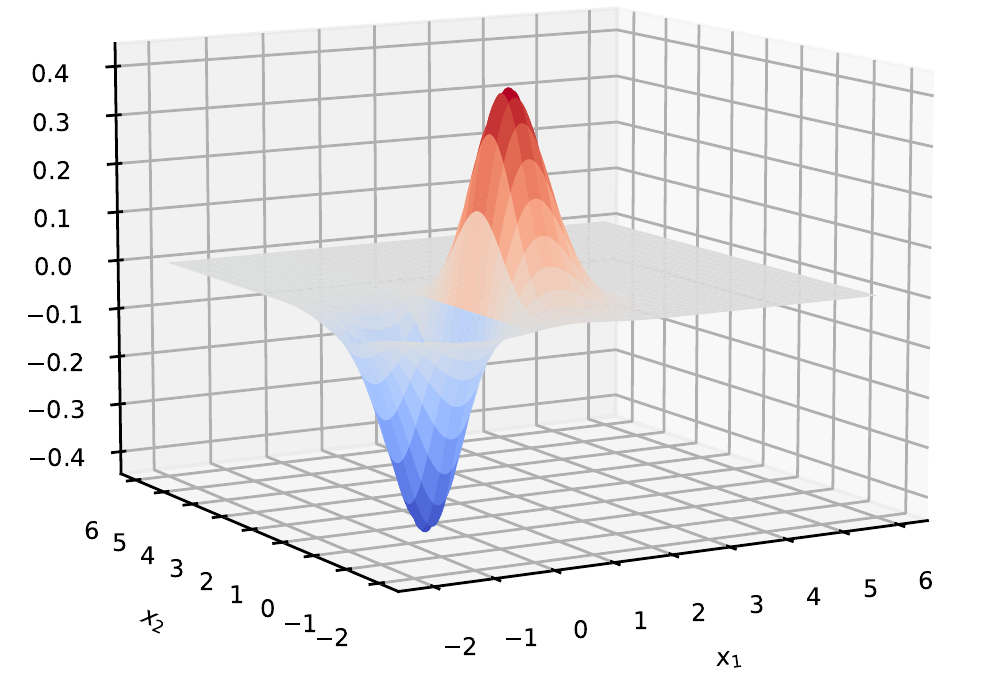}
    \caption{A surface plot of Gramacy \& Lee (2008) function over $[-2,6]^2$.}
    \label{fig:surface_GL2008}
\end{figure}

To demonstrate the model behaviour of GPKSBP, a dataset was generated by the function
\[
f(x_1,x_2) = x_1\exp(-(x_1^2+x_2^2)),
\]
which was used for a similar purpose in \citet{gramacy_gaussian_2008}. Figure \ref{fig:surface_GL2008} plots a 3-D surface of the function over $[-2,6]^2$.
\noindent The function rapidly changes in the region $[-1,1]^2$ and exponentially becomes flat as it moves away from the region. Therefore, it is considered suitable for generating data that reflects non-stationarity.

\begin{figure}[H]
    \centering
    \includegraphics[width=.5\linewidth]{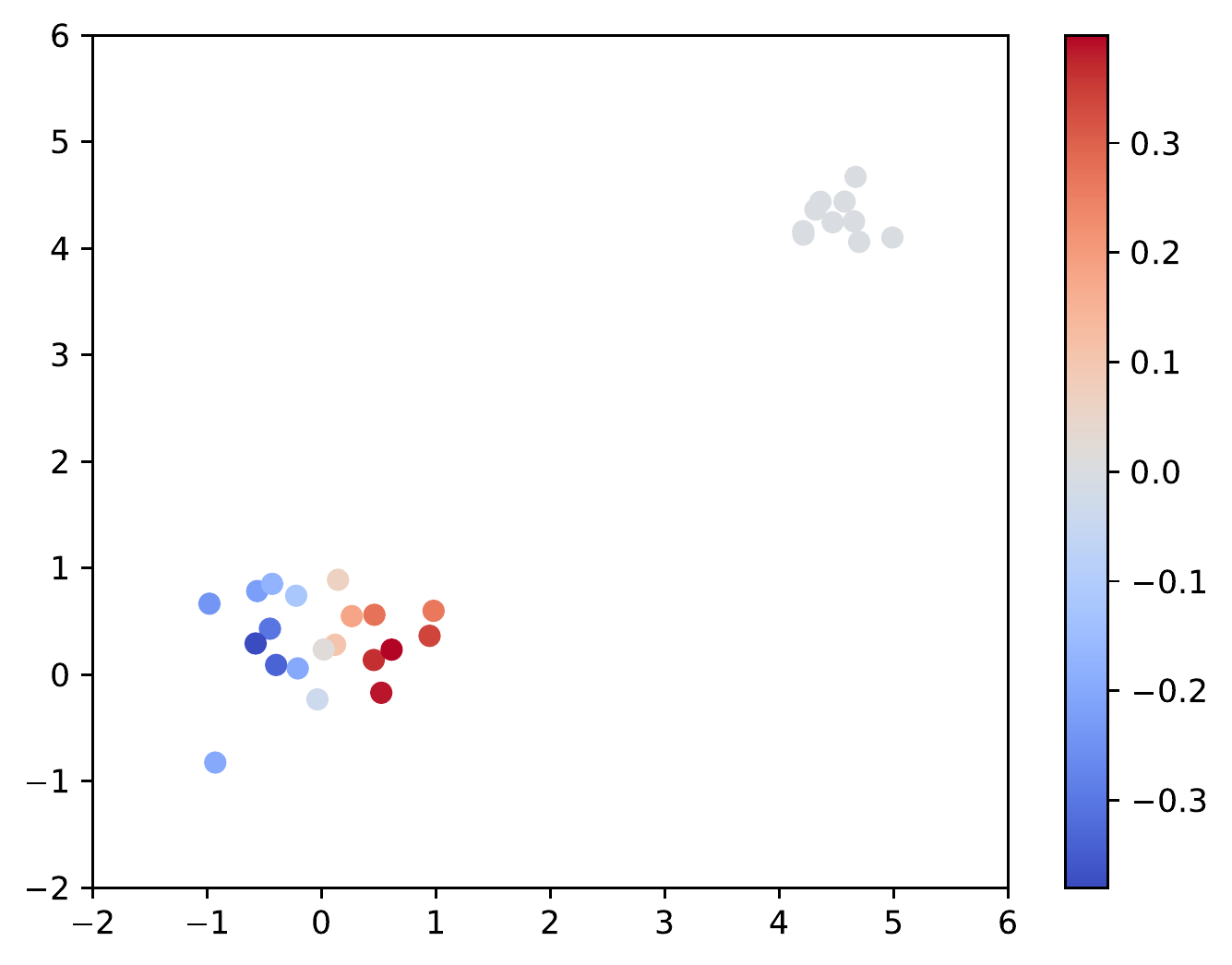}
    \caption{30 sampled locations where Gramacy \& Lee (2008) function were evaluated: 10 from $[-1,0]\times[-1,1]$, 10 from $[0,1]\times[-1,1]$, and 10 from $[4,5]^2$. The corresponding function values are indicated by colours.}
    \label{fig:scatter}
\end{figure}

The dataset was generated by evaluating the function at the following 30 random locations. Given the fact that the global minimum and maximum are located at $(-1/\sqrt{2},0)$ and $(1/\sqrt{2},0)$ respectively, 10 uniform samples were taken from $[-1,0]\times[-1,1]$ and 10 uniform samples were taken from $[0,1]\times[-1,1]$. In addition, to create another cluster distinct from the first cluster, 10 uniform samples were taken from $[4,5]^2$. The realised samples are plotted in Figure \ref{fig:scatter}.

Table \ref{table:GL2008} presents the posterior means of parameters common to all the experts (\(r\), \(\alpha\), \(\beta\)) and parameters specific to the first two experts (\(h\), \(v\), \(\sigma^2\), \(l\)). Note that the results are based on the normalised input \(X\), standardised response \(Y\), and the uniform priors of \(h\) over the unit square \([0,1]^2\). Also note that, given the knowledge of the data-generating process, the noise variance \(\tau^2\) was not learned but fixed at \(10^{-6}\) (not exactly 0 for numerical stability). Each value is the average over over 10,000 MCMC samples after burning the first 10,000 samples.

\begin{table}[H]
\centering
\caption{Estimated posterior means of parameters. The results are based on the standardised response \(Y\) and the normalised input \(X\). \(h\) is also normalised and constrained to \([0,1]^2\). The noise variance \(\tau^2\) was not learned.}
\label{table:GL2008}

\begin{tabular}{c c c}
$r$ & $\alpha$ & $\beta$\\\hline
0.74 & 9.24 & 1.22
\end{tabular}

\begin{tabular}{l l c c c c c}
& Share & $h$ & $v$ & $\sigma^2$ & $l$\\\hline
Expert 1 & 66.5\% & (0.15, 0.19) & 0.96 & 1.88 & (0.19, 0.26)\\\hline
Expert 2 & 33.3\% & (0.83, 0.84) & 0.94 & 0.31 & (1.94, 1.88)
\end{tabular}
\end{table}

On average, virtually only the first two experts were responsible for explaining the whole data---two thirds assigned to Expert 1 and one third assigned to Expert 2. By re-scaling the normalised \(h\) back to the original input space \([-2,6]^2\), we see $(-0.8,-0.48)$ for Expert 1's mean location and $(4.64,4.72)$ for Expert 2's mean location. In other words, located in the steep region, Expert 1 was responsible for explaining the data of rapid change, which is reflected in the small mean length-scale \(l=(0.19,0.26)\). In contrast, located in the flat region, Expert 2 was responsible for explaining the data of little change, which is reflected in the small mean output-scale \(\sigma^2=0.31\) and the large mean length-scale \(l=(1.94,1.88)\).

\begin{figure}[H]
    \centering
    \includegraphics[width=\linewidth]{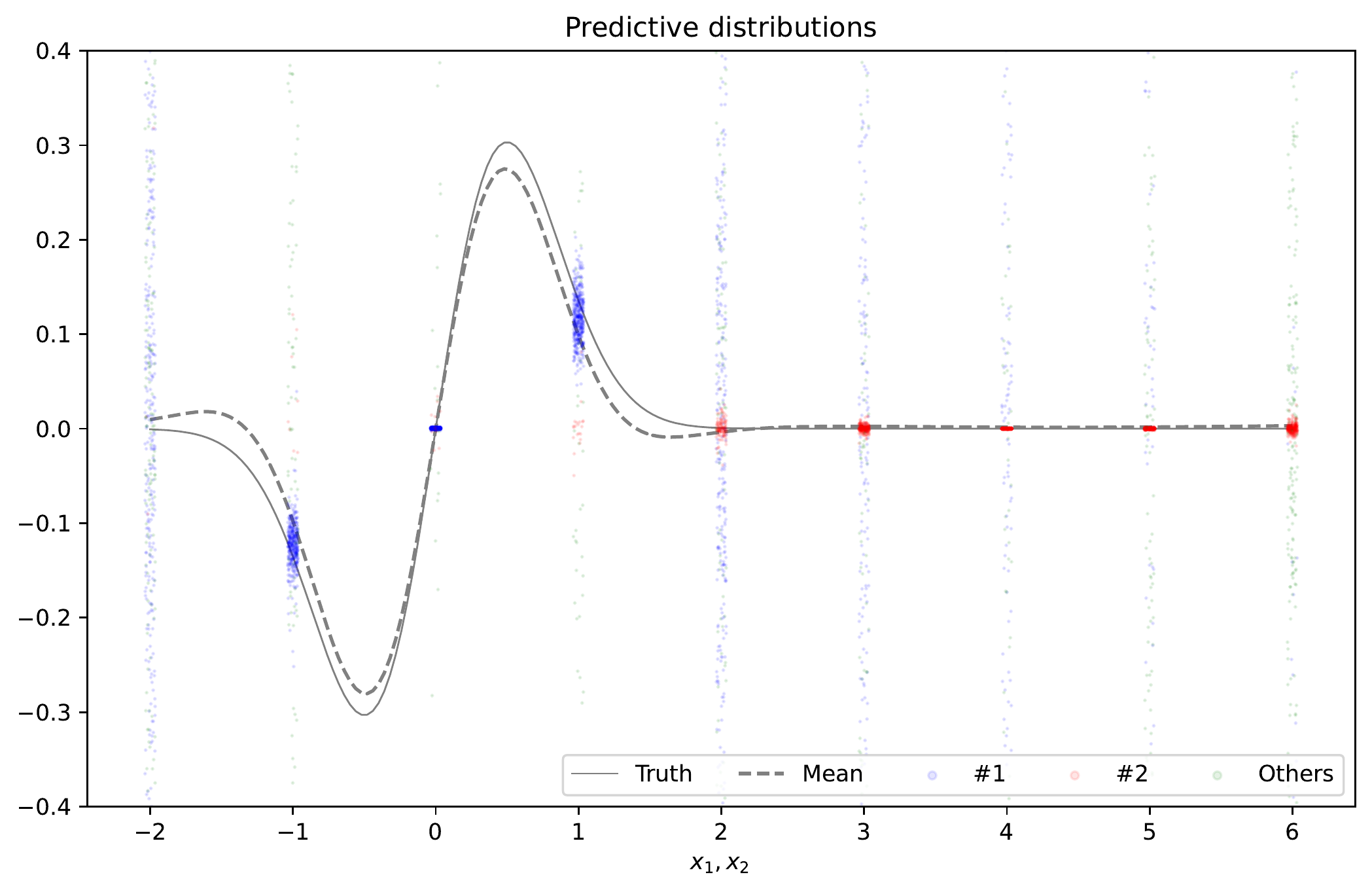}
    \caption{Samples from the posterior predictive distributions at 9 evenly-spaced locations over the diagonal line from $(-2,-2)$ to $(6,6)$, as well as the predictive mean curve. At each of 9 locations, there are 500 samples (jittered for ease of visualisation) in different colours to indicate different realised mixture component GPs. The process used to generate the data is plotted as a thin solid line (Truth).}
    \label{fig:predictive}
\end{figure}

Finally, Figure \ref{fig:predictive} plots samples from the predictive distributions at 9 evenly-spaced locations over the diagonal line from $(-2,-2)$ to $(6,6)$, as well as the predictive mean curve. At each of 9 locations, there are 500 samples (jittered for ease of visualisation) in different colours to indicate different realised mixture component GPs. 500 samples consist of a single sample from each of 500 distinct mixtures of GPs, which results from the remaining MCMC samples after burning the first 10,000 samples from total 20,000 samples and then thinning every 20 samples.

Noticeably, there are some prediction errors on the diagonal line from \((-2,-2)\) to \((-0.5,-0.5)\) and from \((0.3,0.3)\) to \((2,2)\) where the function still changes rapidly. For the first line segment, it is due to the lack of observations over \([-2,-0.5]^2\). For the second line segment, besides the scarcity of data, it is due also to the more even mix of Expert 1 and 2, indicated by the presence of both blue and red samples at \(x=(2,2)\). Since the segment sits between two clusters of data points, the gating function more evenly assigns two experts in this region. While Expert 2 with small \(\sigma^2\) and large \(l\) predicts similar values to those observed in $[4,5]^2$ (i.e., essentially 0) indicated by the mass of red samples, Expert 1 with large \(\sigma^2\) and small \(l\) causes significant prediction uncertainty indicated by the presence of dispersed blue samples.

\subsection{Part 2. Performance comparison}
To reduce the effects of randomness in both dataset generation and posterior sampling, each of five experiments was repeated 30 times using random seeds from 0 to 29, and the average was reported below as the final performance result in that experiment.

\subsubsection{Performance metrics}
The following three performance metrics were used to compare GPKSBP and RG. The performance result at each experiment is the average over all the component results computed against 100 MCMC samples, which were obtained after burning the first half of 20,000 samples and then thinning at every 100 samples. When predicting for test inputs in both GPKSBP and RG, we still need to truncate infinite sequences of GPs and assign the remaining weight to a GP freshly drawn from the priors. However, the remaining weights tend to be very small, and their effects on the performance metrics are negligible.

\begin{itemize}
  \item RMSE (root mean square error) for performance of point-prediction based on the mean of the mixture of GPs.
  \item NLPD (negative log probability density) for performance of distributional prediction based on the mean of predictive densities of a test response value.
  \item CRPS (continuous ranked probability score) for performance of distributional prediction based on the predictive cumulative distribution function of the mixture of GPs. An explicit formula is found in the appendix.
\end{itemize}

\subsubsection{Datasets}
The following are five data generating functions found in the computer simulation literature. Note that function \#5 is made random by additive Gaussian noise \(\mathcal{N}(0,0.05^2)\), as in the source paper \citep{gramacy2009adaptive}. 
\begin{enumerate}
\item Borehole function \citep{worley_deterministic_1987, morris1993bayesian}
  \[f(x) = \frac{2\pi x_3(x_5-x_6)}{\log(x_2/x_1)\left(1+\frac{2x_7x_3}{\log(x_2/x_1)x_1^2x_8} + \frac{x_3}{x_4}\right)},\]
  where
  \begin{gather*}
  x_1\in[0.05,0.15],\; x_2\in[100,50000],\; x_3\in[63070,115600],\; x_4\in[63.1,116]\\
  x_5\in[990,1110],\; x_6\in[700,820],\; x_7\in[1120,1680],\; x_8\in[9855,12045].
  \end{gather*}
\item Dette \& Pepelyshev exponential function \citep{dette2010generalized}
  \[f(x) = 100\left(\exp(-2/x_1^{1.75}) + \exp(-2/x_2^{1.5}) + \exp(-2/x_3^{1.25})\right)\]
  where \(x\in[0,1]^3\).
\item Dette \& Pepelyshev 8-D function \citep{dette2010generalized}
  \[f(x) = 4(x_1-2+8x_2-8x_2^2)^2 + (3-4x_2)^2 + 16\sqrt{x_3+1}(2x_3-1)^2 + \sum_{i=4}^8i\log\left(1+\sum_{j=3}^ix_j\right)\]
  where \(x\in[0,1]^8\).
\item Franke function \citep{franke1979critical,haaland2011accurate}
  \begin{align*}
  f(x) &= 0.75\exp\left(-\frac{(9x_1-2)^2}{4}-\frac{(9x_2-2)^2}{4}\right)
          +0.75\exp\left(-\frac{(9x_1+1)^2}{49}-\frac{(9x_2+1)^2}{10}\right)\\
       &\quad\quad +0.5\exp\left(-\frac{(9x_1-7)^2}{4}-\frac{(9x_2-3)^2}{4}\right)
                   -0.2\exp\left(-(9x_1-4)^2-(9x_2-7)^2\right),
  \end{align*}
  where \(x\in[0,1]^2\).
\item Gramacy \& Lee function \citep{gramacy2009adaptive}
  \[f(x) = \exp\left(\sin((0.9(x_1+0.48))^{10})\right) + x_2x_3 + x_4 + \epsilon\]
  where  \(\epsilon \sim \mathcal{N}(0,0.05^2)\) and \(x\in[0,1]^6\) with inactive \(x_5,x_6\).
\end{enumerate}

In each experiment, we drew uniform samples \(x\) from the function domain and evaluated them to obtain the corresponding \(y\): 30 samples for the training dataset and 300 samples for the test dataset. The training sample size \(N=30\) was chosen to be relatively small for two practical reasons. First, when emulators are used, it is often the case that simulations are very expensive and time-consuming. Therefore, it is reasonable to restrict the sample size and investigate small-sample predictive performance. Second, we would like to reduce the chance of immature convergence in the posterior sampling and chose 20,000 MCMC samples. To repeat 30 times (due to 30 random seeds) each of five different experiments for both methods, we needed to restrict the sample size.

\subsubsection{Results}
Table \ref{table:performance} presents results. It turned out that GPKSBP outperformed RG in terms of all three metrics in all of five experiments except RMSE for dataset 1.

\begin{table}[H]
\centering
\caption{Performance results from five experiments. At each pair, a smaller value indicates better performance. Note that RMSE and CRPS are in the unit of response variables, which are standardised in all experiments but may slightly differ across the datasets.}
\label{table:performance}
\vspace{0.3cm}
\small
\begin{tabular}{l |r r|r r|r r|r r|r r}
 & \multicolumn{2}{l|}{dataset 1} & \multicolumn{2}{l|}{dataset 2} & \multicolumn{2}{l|}{dataset 3} & \multicolumn{2}{l|}{dataset 4} & \multicolumn{2}{l}{dataset 5}\\
 & GPKSBP & RG & GPKSBP & RG & GPKSBP & RG & GPKSBP & RG & GPKSBP & RG\\\hline
RMSE & 0.29 & 0.25 & 0.23 & 0.32 & 0.73 & 0.91 & 0.26 & 0.30 & 0.96 & 1.00\\\hline
NLPD & -0.24 & -0.19 & -0.63 & -0.23 & 0.99 & 1.31 & -0.43 & -0.03 & 1.37 & 1.50\\\hline
CRPS & 0.12 & 0.13 & 0.07 & 0.15 & 0.38 & 0.51 & 0.11 & 0.14 & 0.53 & 0.57
\end{tabular}
\end{table}

\begin{figure}[H]
    \centering
    \includegraphics[width=.6\linewidth]{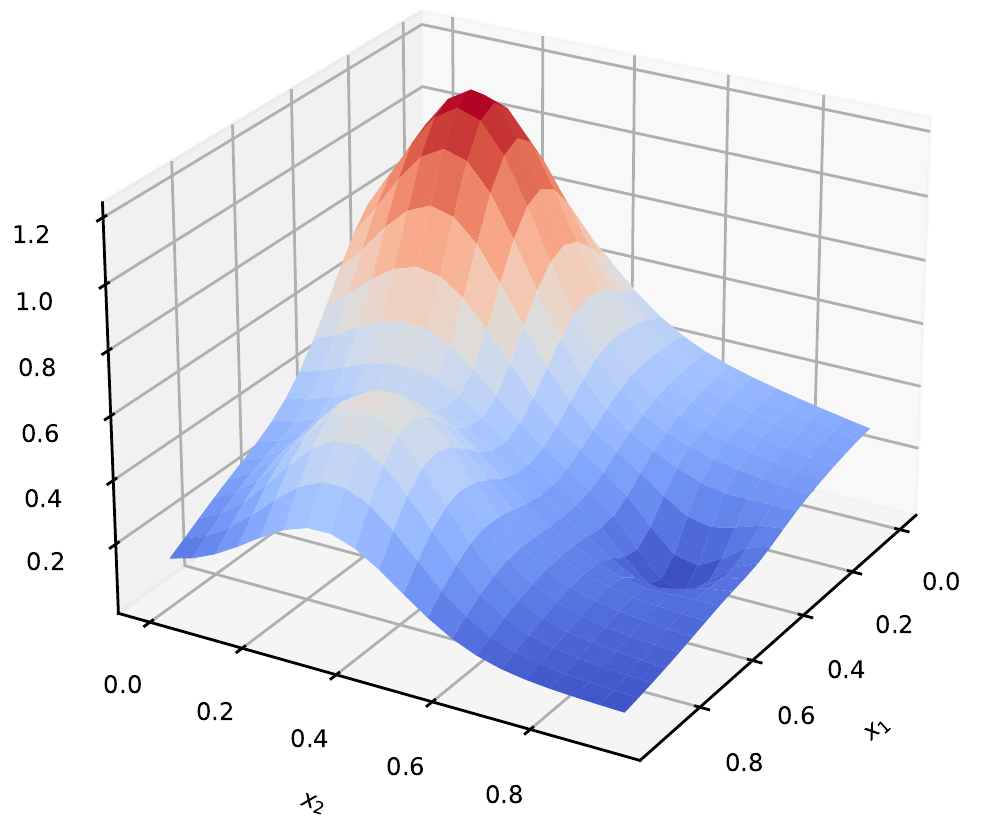}
    \caption{A surface plot of Franke function (dataset 4). Notice that the axes are rotated for improved visibility of key features.}
    \label{fig:surface_Franke}
\end{figure}

\noindent To gain some insight into the cause of performance difference, we examine in detail one of the results from dataset 4 generated by Franke function, which is defined on \([0,1]^2\) and simple to visualise. As indicated by similar blue colours in Figure \ref{fig:surface_Franke}, a region over \(x_2\geq 0.6\) is relatively flat.

\noindent The following tables contain the estimated posterior means of parameters of GPKSBP and RG under random seed 0, one of 30 seeds used to produce the final results reported above.

\begin{table}[H]
\centering
\caption{Estimated posterior means of parameters by GPKSBP for dataset 4 using random seed 0.}
\label{table:GPKSBP_Franke}

\begin{tabular}{c c c}
$r$ & $\alpha$ & $\beta$\\\hline
1.78 & 7.18 & 1.54
\end{tabular}

\begin{tabular}{l l c c c c c}
& Share & $h$ & $v$ & $\sigma^2$ & $l$\\\hline
Expert 1 & 73.0\% & (0.44, 0.30) & 0.85 & 1.62 & (0.42, 0.36)\\\hline
Expert 2 & 22.8\% & (0.37, 0.51) & 0.88 & 2.19 & (1.14, 1.04)
\end{tabular}

\caption{Estimated posterior means of parameters by RG for dataset 4 using random seed 0. Note that \(\alpha\), \(h\), and \(v\) are applicable only to GPKSBP.}
\label{table:RG_Franke}

\begin{tabular}{c c c}
$r$ & $\beta$\\\hline
0.22 & 1.65
\end{tabular}

\begin{tabular}{l c c c c c}
& Share & $\sigma^2$ & $l$\\\hline
Expert 1 & 49.4\% & 2.69 & (0.88,0.72)\\\hline
Expert 2 & 20.4\% & 2.77 & (1.05,0.98)\\\hline
Expert 3 & 12.4\% & 3.01 & (1.06,0.88)\\\hline
Expert 4 & 7.6\% & 3.12 & (1.06,0.91)\\\hline
Expert 5 & 4.5\% & 3.4 & (1.00,0.91)
\end{tabular}
\end{table}

\noindent On average, GPKSBP relies on two experts to explain 95\% of the data points. In contrast, RG relies on five experts to explain 95\% of the data points. However, these experts are not well distinguished from each other but share similar values of \(\sigma^2\) and \(l\). It turned out that this was a common feature of RG observed in many other experiments. If GPs are not uniquely characterised to explain local regions, they are not ``experts'' in the spirit of mixture-of-expert models. Unnecessary split of scarce data into similar GPs merely reduces the amount of data on which each prediction is based. It seems that over-split of the gating function is a key factor that reduces RG's predictive performance.

\section{Discussion}\label{sec:discussion}
We constructed the new mixture model of GP experts (GPKSBP), designed the posterior sampler utilising the slice sampling, and demonstrated its improved predictive performance against the baseline model (RG). Due to the significance of the baseline model, which defines this class of models (mixtures of GP experts based on dependent Dirichlet processes), we focused on the improvement in predictive performance relative to the baseline. In particular, we highlighted the issue that RG's gating function tends to over-split the input space, thereby reducing its predictive performance due to the fewer data points each expert bases its prediction on. When applying GPKSBP to specific problems, fine-tuning will be possible and in fact recommended. In what follows, we discuss several aspects for fine-tuning GPKSBP.

Among many possible dependent Dirichlet processes that can be used to design gating functions, our choice of KSBP is due to the continuity between RG and GPKSBP. Considering the role of gating functions, which is to probabilistically assign each data point to an expert, RG uses the kernel based on Euclidean distance to measure the closeness of input \(x\) to the data points assigned to each of the candidate experts. We find this kernel having intuitive appeal for model interpretation. Therefore, we use the same kernel to measure the closeness of input \(x\) directly to each of the candidate experts using their locations \(h\) in the input space. While intuitive and increasing the model interpretability, it is by no means the best choice in all applications with different modelling goals. For example, in classification problems, it could be more sensible to use a metric that returns 1 if the distance is within a threshold and 0 otherwise, of which some theoretical properties have been proved \citep{dunson_kernel_2008}.

Similarly, we adopt the squared exponential covariance function simply because it is used for RG. Even within the class of stationary covariance functions, many choices are available and some are more suitable for particular applications than others. For example, in practice, Mat\'{e}rn covariance functions are popular because they are generalisation of the squared exponential and allow practitioners to control the smoothness of GPs.

Our specifications of priors is due to the convenience for design/implementation of the posterior sampler and the fact that we try to avoid undue tuning of GPKSBP to the datasets used in the experiments. While the beta prior for \(v\) is inherently tied to Dirichlet processes and more generally stick-breaking processes, the other priors can be chosen to suit each application. For example, in geo-statistical applications, there may be some useful information about expert locations based on geographical and/or geological features. If so, the uniform prior for \(h\) should be adapted accordingly. Moreover, even within the same family of distributions, their own parameters can be adapted for suitable prior means and variances. For example, if domain knowledge suggests noisy underlying processes, the prior for \(\tau_i^2\) may be chosen to have larger mean than 1 implied by gamma\((2,0.5)\) used in the experiments.

While mixtures of GP experts simultaneously address both of the inherent limitations of standard GPs, scalability and predictive performance, there exists a trade-off between two benefits. Intuitively, as the number of active experts increases, scalability increases but predictive performance may decrease as demonstrated by the over-split of RG in the experiments. Thus, modellers must be aware of the trade-off and, in some cases, explicitly control the trade-off to suit particular applications under specific constraints on computational resources. In GPKSBP, one way to control the trade-off is through the priors for \(r\), \(\alpha\), and \(\beta\), which directly govern the underlying dependent Dirichlet process that in turn dictates the gating function. Under the kernel based on scaled Euclidean distance, as \(r\) becomes small, input \(x\) becomes more distant from the existing experts, each of which is partially characterised by location \(h_i\), and therefore the gating function more likely assigns \(x\) to a new expert. The role of \(\alpha\) and \(\beta\) is to control the stick-breaking probabilities \(v\) through the beta prior distribution; the smaller \(v_i\), the more likely the gating function assigns \(x\) to the \(i+1\)th expert. Of course, since these parameters are intimately coupled with the other components, modellers must strive for holistic tuning of the overall model behaviour.

\section*{Appendix}
\subsection*{HMC specifications}
Here we explain the implementation details of the HMC updates in our algorithm. HMC is used to update \( \sigma_i^2, l_i, \tau_i^2\)  and $r$. Note that these parameters are all positive.
Suppose we are interested in sampling a positive parameter \(\theta\). Because HMC works better on unconstrained spaces, we sample $\log(\theta)$ instead. It is straightforward to do so. Let $w = g(\theta) = \log(\theta)$. Then
$$p_w(w) = p_\theta(g^{-1}(w))\left\vert\frac{\partial}{\partial w}g^{-1}(w)\right\vert = p_\theta(e^w)e^w.$$
Using this change of variable technique we can obtain priors for $\log(\theta)$ that correspond to the prior we choose for $\theta$ and rewrite the potential energy $U$ as a function of $w$. The HMC update then proceeds as in \citet{neal2011mcmc}, with the leapfrog integrator making updates for $\log(\theta)$ instead of $\theta$.

In implementing the HMC step, the step size for each expert is chosen using the dual averaging algorithm in \cite{hoffman2014no}. This algorithm adaptively changes the step size during the burn-in period to achieve a desired acceptance rate, which we set at 0.8 in our experiments. For our problem, to prevent divergent transitions, the step size is capped at 0.05. The number of leapfrog step is set at 5.

\subsection*{Formula for continuous ranked probability score}
Since GPs make distributional prediction, CRPS is considered as a suitable performance metric \citep{gneiting_strictly_2007}. When predictive distribution is a Gaussian mixture, a closed-form expression for CRPS is available \citep{grimit_continuous_2006}. Let \(f(x)\) be a Gaussian mixture random variable induced by the predictive distribution at test location \(x\), and \(y\) be the corresponding response value. Then,
\[
CRPS(f(x), y) = \sum_{i=1}^{i^*+1} w_i\Psi(y-\mu_i,\sigma_i^2) - \frac{1}{2}\sum_{i=1}^{i^*+1}\sum_{j=1}^{i^*+1}w_iw_j\Psi(\mu_i-\mu_j, \sigma_i^2+\sigma_j^2),
\]
where
\[
\Psi(\mu,\sigma^2) = 2\sigma\phi\Big(\frac{\mu}{\sigma}\Big) + \mu\left[2\Phi\Big(\frac{\mu}{\sigma}\right)-1\Big],
\]
and \(\phi\) and \(\Phi\) denote respectively the PDF and the CDF of the standard normal distribution. Note that each summation is over \(i^*+1\) components because, at each MCMC iteration, the mixture is truncated at the \(i^*\)th GP by the slice sampling and, at prediction, a new GP is freshly drawn from the prior for the remaining weight.

\subsection*{Rank-1 update of the inverse covariance matrix}
A change in \(s_n\) after sampling induces a change in the data points associated with both the previous expert and the new expert, i.e. a change in the inverse covariance matrices, which may have impact on the likelihood for \(n+1, n+2, \dots\). Since naive updates of these matrices are computationally expensive, \citet{rasmussen_infinite_2001} suggest use of rank-1 updates when inverting a covariance matrix, which has only a single row and column added/removed from the existing covariance matrix whose inverse is already computed and stored.

Let \(K\) be a \(N \times N\) covariance matrix. Also, let sub-matrices of \(K\) and its inverse be denoted as follows.
\[
K =
  \begin{bmatrix}
  A & B\\
  C & D\\
  \end{bmatrix}
,\quad
K^{-1} =
  \begin{bmatrix}
  a & b\\
  c & d\\
  \end{bmatrix}.
\]
By definition,
\[
I = KK^{-1} =
  \begin{bmatrix}
  Aa+Bc & Ab+Bd\\
  Ca+Dc & Cb+Dd
  \end{bmatrix}
\]
where
\[
Ab+Bd = Ca+Dc = 0.
\]
\(Aa+Bc\) and \(Cb+Dd\) are identity matrices of conformable size. The goal of rank-1 update is to compute either
\begin{enumerate}
\item a \(N-1 \times N-1\) inverse matrix \(D^{-1}\) with the knowledge of \(K^{-1}\), or
\item a \(N+1 \times N+1\) inverse matrix \(K^{-1}\) with the knowledge of \(K\) and \(D^{-1}\).
\end{enumerate}

Case \#1.
\[
D^{-1} = d - \frac{cb}{a}.
\]
Note that \(a\) is a scalar.

Case \#2.
\begin{align*}
  a &= \frac{1}{A - BD^{-1}C}\\
  c &= -D^{-1}Ca\\
  b &= c^T\\
  d &= D^{-1} - D^{-1}Cb.
\end{align*}

\subsection*{Posterior sampler of \(r\) in Rasmussen \& Ghahramani (2001)}
Samples are taken from the pseudo-posterior, which is obtained by multiplying a prior and the pseudo-likelihood given \(\{s_n\}\). The latter is
\begin{align*}
  p(r|\{s_n\}) &= \prod_n p(s_n|s_{-n},r)\\
               &=
               \begin{cases}
                   \frac{N_{-n,i}}{N-1+\beta} & \text{if } i\in\{1,\dots,i^*\}\\
                   \frac{\beta}{N-1+\beta} & \text{otherwise}\\
               \end{cases}
\end{align*}
where \(N_{-n,i}\) denote the occupation number of expert \(i\in\{1,\dots,i^*\}\) excluding the \(n\)th data point. \(N_{-n,i}\) is made input-dependent based on the following formula:
\[
N_{-n,i} = (N-1)\frac{\sum_{n'\neq n}\kappa(x_n,x_{n'})\delta_{i,s_{n'}}}{\sum_{n'\neq n}\kappa(x_n,x_{n'})}.
\]
For HMC, the gradient of the log-pseudo-likelihood is computed as follows.
\begin{align*}
\nabla \log p(r|\{s_n\})
&= \sum_n \nabla\log(p(s_n|s_{-n},r))\\
&= \sum_n \nabla\log(N_{-n,i})\\
&= \sum_n\left[\nabla\log\left(\sum_{n'\neq n}\kappa(x_n,x_{n'})\delta_{i,s_{n'}}\right) - \nabla\log\left(\sum_{n'\neq n}\kappa(x_n,x_{n'})\right)\right]\\
&= \sum_n\left[\frac{\sum_{n'\neq n}\nabla\kappa(x_n,x_{n'})\delta_{i,s_{n'}}}{\sum_{n'\neq n}\kappa(x_n,x_{n'})\delta_{i,s_{n'}}} - \frac{\sum_{n'\neq n}\nabla\kappa(x_n,x_{n'})}{\sum_{n'\neq n}\kappa(x_n,x_{n'})}\right],
\end{align*}
where each derivative is
\[
\frac{d}{dr}\kappa(x_n,x_{n'}) = 2\kappa(x_n,x_{n'})\frac{\|x_n-x_{n'}\|^2}{r^3}.
\]

\subsection*{Posterior of \(\beta\) under the conjugacy}
Below follows from \citet[20]{muller_bayesian_2015}. Suppose the standard Dirichlet process \(\text{DP}(\beta, G_0)\) with prior \(\beta \sim \text{gamma}(a,b)\). First, sample auxiliary \(\phi \sim \text{beta}(\beta+1, N)\) using the current value of \(\beta\) where \(N\) is the number of data points. Then,

\begin{align*}
odds &= \frac{a+|e|-1}{N(b-\log(\phi))}\\
q &= \frac{odds}{1+odds}\\
\beta | \phi, |e| &\sim
\begin{cases}
  \text{gamma}(a + |e|,\; b-\log(\phi)) & \text{w.p. } q\\
  \text{gamma}(a + |e| - 1,\; b-\log(\phi)) & \text{w.p. } 1-q
\end{cases}
\end{align*}
where \(|e|\) is the number of experts with at least one data point assigned.


\printbibliography

@inproceedings{snoek_input_2014,
	location = {Bejing, China},
	title = {Input Warping for Bayesian Optimization of Non-Stationary Functions},
	volume = {32},
	url = {https://proceedings.mlr.press/v32/snoek14.html},
	series = {Proceedings of Machine Learning Research},
	pages = {1674--1682},
	booktitle = {Proceedings of the 31st International Conference on Machine Learning},
	publisher = {{PMLR}},
	author = {Snoek, Jasper and Swersky, Kevin and Zemel, Rich and Adams, Ryan},
	editor = {Xing, Eric P. and Jebara, Tony},
	date = {2014-06-22},
	note = {Issue: 2},
}

@article{nott_estimation_2002,
	title = {Estimation of nonstationary spatial covariance structure},
	volume = {89},
	pages = {819--829},
	number = {4},
	journaltitle = {Biometrika},
	author = {Nott, David J and Dunsmuir, William {TM}},
	date = {2002},
	note = {Publisher: Oxford University Press},
	}

@article{rasmussen_infinite_2001,
	title = {Infinite mixtures of Gaussian process experts},
	volume = {14},
	journaltitle = {Advances in neural information processing systems},
	author = {Rasmussen, Carl and Ghahramani, Zoubin},
	date = {2001},
	}

@article{kim_analyzing_2005,
	title = {Analyzing nonstationary spatial data using piecewise Gaussian processes},
	volume = {100},
	pages = {653--668},
	number = {470},
	journaltitle = {Journal of the American Statistical Association},
	author = {Kim, Hyoung-Moon and Mallick, Bani K and Holmes, Chris C},
	date = {2005},
	note = {Publisher: Taylor \& Francis},
	}

@article{neal_markov_2000,
	title = {Markov chain sampling methods for Dirichlet process mixture models},
	volume = {9},
	pages = {249--265},
	number = {2},
	journaltitle = {Journal of computational and graphical statistics},
	author = {Neal, Radford M},
	date = {2000},
	note = {Publisher: Taylor \& Francis},
	}

@article{quintana_dependent_2022,
	title = {The Dependent Dirichlet Process and Related Models},
	volume = {37},
	url = {https://doi.org/10.1214/20-STS819},
	doi = {10.1214/20-STS819},
	pages = {24 -- 41},
	number = {1},
	journaltitle = {Statistical Science},
	author = {Quintana, Fernando A. and Müller, Peter and Jara, Alejandro and {MacEachern}, Steven N.},
	date = {2022},
	note = {Publisher: Institute of Mathematical Statistics},
	}

@article{schmidt_bayesian_2003,
	title = {Bayesian inference for non-stationary spatial covariance structure via spatial deformations},
	volume = {65},
	doi = {https://doi.org/10.1111/1467-9868.00413},
	pages = {743--758},
	number = {3},
	journaltitle = {Journal of the Royal Statistical Society: Series B (Statistical Methodology)},
	author = {Schmidt, Alexandra M. and O'Hagan, Anthony},
	date = {2003},
	}

@article{dunson_kernel_2008,
	title = {Kernel stick-breaking processes},
	volume = {95},
	issn = {0006-3444},
	url = {https://doi.org/10.1093/biomet/asn012},
	doi = {10.1093/biomet/asn012},
	pages = {307--323},
	number = {2},
	journaltitle = {Biometrika},
	author = {Dunson, David B. and Park, Ju-Hyun},
	date = {2008-04},
	}

@article{papaspiliopoulos_retrospective_2008,
	title = {Retrospective Markov chain Monte Carlo methods for Dirichlet process hierarchical models},
	volume = {95},
	pages = {169--186},
	number = {1},
	journaltitle = {Biometrika},
	author = {Papaspiliopoulos, Omiros and Roberts, Gareth O},
	date = {2008},
	note = {Publisher: Oxford University Press},
	}

@article{kalli_slice_2011,
	title = {Slice sampling mixture models},
	volume = {21},
	issn = {1573-1375},
	url = {https://doi.org/10.1007/s11222-009-9150-y},
	doi = {10.1007/s11222-009-9150-y},
	pages = {93--105},
	number = {1},
	journaltitle = {Statistics and Computing},
	shortjournal = {Statistics and Computing},
	author = {Kalli, Maria and Griffin, Jim E. and Walker, Stephen G.},
	date = {2011-01-01},
	}

@article{walker_sampling_2007,
	title = {Sampling the Dirichlet Mixture Model with Slices},
	volume = {36},
	doi = {10.1080/03610910601096262},
	pages = {45--54},
	number = {1},
	journaltitle = {Communications in Statistics - Simulation and Computation},
	author = {Walker, Stephen G.},
	date = {2007},
	}

@article{gramacy_gaussian_2008,
	title = {Gaussian processes and limiting linear models},
	volume = {53},
	issn = {0167-9473},
	url = {https://www.sciencedirect.com/science/article/pii/S0167947308003307},
	doi = {https://doi.org/10.1016/j.csda.2008.06.020},
	pages = {123--136},
	number = {1},
	journaltitle = {Computational Statistics \& Data Analysis},
	author = {Gramacy, Robert B. and Lee, Herbert K. H.},
	date = {2008},
	}

@book{muller_bayesian_2015,
	title = {Bayesian nonparametric data analysis},
	volume = {1},
	publisher = {Springer},
	author = {Müller, Peter and Quintana, Fernando Andrés and Jara, Alejandro and Hanson, Tim},
	date = {2015},
}

@article{hoffman2014no,
  title={The No-U-Turn sampler: adaptively setting path lengths in Hamiltonian Monte Carlo.},
  author={Hoffman, Matthew D and Gelman, Andrew and others},
  journal={J. Mach. Learn. Res.},
  volume={15},
  number={1},
  pages={1593--1623},
  year={2014}
}

@article{neal2011mcmc,
	title={{MCMC} using {H}amiltonian dynamics},
	author={Neal, Radford M.},
	journal={Handbook of Markov Chain Monte Carlo},
	volume={2},
	pages={113--162},
	year={2011}
}

@article{wang2012gaussian,
  title={Gaussian process regression with heteroscedastic or non-Gaussian residuals},
  author={Wang, Chunyi and Neal, Radford M},
  journal={arXiv preprint arXiv:1212.6246},
  year={2012}
}

@article{jacobs1991adaptive,
  title={Adaptive mixtures of local experts},
  author={Jacobs, Robert A and Jordan, Michael I and Nowlan, Steven J and Hinton, Geoffrey E},
  journal={Neural computation},
  volume={3},
  number={1},
  pages={79--87},
  year={1991},
  publisher={MIT Press}
}

@article{liu2020gaussian,
  title={When Gaussian process meets big data: A review of scalable GPs},
  author={Liu, Haitao and Ong, Yew-Soon and Shen, Xiaobo and Cai, Jianfei},
  journal={IEEE transactions on neural networks and learning systems},
  volume={31},
  number={11},
  pages={4405--4423},
  year={2020},
  publisher={IEEE}
}

@book{rasmussen2006gaussian,
  title={Gaussian processes for machine learning},
  author={Rasmussen, Carl Edward and Williams, Christopher KI},
  volume={2},
  number={3},
  year={2006},
  publisher={MIT press Cambridge, MA}
}

@article{anderes2008estimating,
  title={Estimating deformations of isotropic Gaussian random fields on the plane},
  author={Anderes, Ethan B and Stein, Michael L},
  journal={The Annals of Statistics},
  volume={36},
  number={2},
  pages={719--741},
  year={2008},
  publisher={Institute of Mathematical Statistics}
}

@article{paciorek2006spatial,
  title={Spatial modelling using a new class of nonstationary covariance functions},
  author={Paciorek, Christopher J and Schervish, Mark J},
  journal={Environmetrics: The official journal of the International Environmetrics Society},
  volume={17},
  number={5},
  pages={483--506},
  year={2006},
  publisher={Wiley Online Library}
}

@book{santner_design_2003,
	title = {The design and analysis of computer experiments},
	volume = {1},
	publisher = {Springer},
	author = {Santner, Thomas J and Williams, Brian J and Notz, William I and Williams, Brain J},
	date = {2003},
}

@book{cressie_statistics_1993,
	title = {Statistics for spatial data},
	publisher = {Wiley-Interscience},
	author = {Cressie, Noel},
	date = {1993},
}

@article{shahriari_taking_2015,
	title = {Taking the human out of the loop: A review of bayesian optimization},
	volume = {104},
	issn = {0018-9219},
	pages = {148--175},
	number = {1},
	journaltitle = {Proceedings of the {IEEE}},
	shortjournal = {Proceedings of the {IEEE}},
	author = {Shahriari, Bobak and Swersky, Kevin and Wang, Ziyu and Adams, Ryan P and De Freitas, Nando},
	date = {2015},
}

@article{fuglstad_does_2015,
	title = {Does non-stationary spatial data always require non-stationary random fields?},
	volume = {14},
	issn = {2211-6753},
	url = {https://www.sciencedirect.com/science/article/pii/S2211675315000780},
	doi = {10.1016/j.spasta.2015.10.001},
	pages = {505--531},
	journaltitle = {Spatial Statistics},
	shortjournal = {Spatial Statistics},
	author = {Fuglstad, Geir-Arne and Simpson, Daniel and Lindgren, Finn and Rue, Håvard},
	date = {2015-11-01},
}

@report{worley_deterministic_1987,
	title = {Deterministic uncertainty analysis},
	institution = {Oak Ridge National Lab., {TN} ({USA})},
	author = {Worley, Brian A},
	date = {1987},
}

@article{gneiting_strictly_2007,
	title = {Strictly Proper Scoring Rules, Prediction, and Estimation},
	volume = {102},
	issn = {0162-1459},
	url = {https://doi.org/10.1198/016214506000001437},
	doi = {10.1198/016214506000001437},
	pages = {359--378},
	number = {477},
	journaltitle = {Journal of the American Statistical Association},
	shortjournal = {Journal of the American Statistical Association},
	author = {Gneiting, Tilmann and Raftery, Adrian E},
	date = {2007-03-01},
	note = {Publisher: Taylor \& Francis},
}

@article{grimit_continuous_2006,
	title = {The continuous ranked probability score for circular variables and its application to mesoscale forecast ensemble verification},
	volume = {132},
	pages = {2925--2942},
	number = {621},
	journaltitle = {Quarterly Journal of the Royal Meteorological Society: A journal of the atmospheric sciences, applied meteorology and physical oceanography},
	author = {Grimit, Eric P and Gneiting, Tilmann and Berrocal, Veronica J and Johnson, Nicholas A},
	date = {2006},
	note = {Publisher: Wiley Online Library},
}

@article{dette2010generalized,
  title={Generalized Latin hypercube design for computer experiments},
  author={Dette, Holger and Pepelyshev, Andrey},
  journal={Technometrics},
  volume={52},
  number={4},
  pages={421--429},
  year={2010},
  publisher={Taylor \& Francis}
}

@article{gramacy2009adaptive,
  title={Adaptive design and analysis of supercomputer experiments},
  author={Gramacy, Robert B and Lee, Herbert KH},
  journal={Technometrics},
  volume={51},
  number={2},
  pages={130--145},
  year={2009},
  publisher={Taylor \& Francis}
}

@article{escobar1995bayesian,
  title={Bayesian density estimation and inference using mixtures},
  author={Escobar, Michael D and West, Mike},
  journal={Journal of the american statistical association},
  volume={90},
  number={430},
  pages={577--588},
  year={1995},
  publisher={Taylor \& Francis}
}

@techreport{franke1979critical,
  title={A critical comparison of some methods for interpolation of scattered data},
  author={Franke, Richard},
  year={1979},
  institution={Naval Postgraduate School Monterey CA}
}

@article{haaland2011accurate,
  title={Accurate emulators for large-scale computer experiments},
  author={Haaland, Ben and Qian, Peter ZG},
  year={2011}
}

@article{morris1993bayesian,
  title={Bayesian design and analysis of computer experiments: use of derivatives in surface prediction},
  author={Morris, Max D and Mitchell, Toby J and Ylvisaker, Donald},
  journal={Technometrics},
  volume={35},
  number={3},
  pages={243--255},
  year={1993},
  publisher={Taylor \& Francis}
}

@article{gilks1992adaptive,
  title={Adaptive rejection sampling for Gibbs sampling},
  author={Gilks, Walter R and Wild, Pascal},
  journal={Journal of the Royal Statistical Society: Series C (Applied Statistics)},
  volume={41},
  number={2},
  pages={337--348},
  year={1992},
  publisher={Wiley Online Library}
}

\end{document}